%% file: root.tex
\pdfoutput=1
\documentclass[letterpaper, 10 pt, conference]{ieeeconf}  

\IEEEoverridecommandlockouts                             
\usepackage{graphicx} 
\usepackage{amsmath} 
\usepackage{amssymb}  
\usepackage{booktabs}
\usepackage{wrapfig}
\usepackage{hyperref}

\let\labelindent\relax
\usepackage{enumitem}

\newcommand{\proposedmodel}{MagNet}

\title{\LARGE \bf
\proposedmodel: Discovering Multi-agent Interaction Dynamics using Neural Network
}

\author{Priyabrata Saha, Arslan Ali, Burhan A. Mudassar, Yun Long and Saibal Mukhopadhyay
\thanks{Authors are with School of Electrical and Computer Engineering,
Georgia Tech , Atlanta, USA.
{\tt\small \{priyabratasaha, arslanali, burhan.mudassar, yunlong\}@gatech.edu}, {\tt\small saibal.mukhopadhyay@ece.gatech.edu}. \newline 
This material is based on work sponsored by the Army Research Office and was accomplished under Grant Number W911NF-19-1-0447. The views and conclusions contained in this document are those of the authors and should not be interpreted as representing the official policies, either expressed or implied, of the Army Research Office or the U.S. Government.}
}

\begin{document}

\maketitle
\thispagestyle{empty}
\pagestyle{empty}

%%%%%%%%%%%%%%%%%%%%%%%%%%%%%%%%%%%%%%%%%%%%%%%%%%%%%%%%%%%%%%%%%%%%%%%%%%%%%%%%
\begin{abstract}
We present the~\proposedmodel, a neural network-based multi-agent interaction model to discover the governing dynamics and predict evolution of a complex multi-agent system from observations. We formulate a multi-agent system as a coupled non-linear network with a generic ordinary differential equation (ODE) based state evolution, and develop a neural network-based realization of its time-discretized model.~\proposedmodel\ is trained to discover the core dynamics of a multi-agent system from observations, and tuned on-line to learn agent-specific parameters of the dynamics to ensure accurate prediction even when physical or relational attributes of agents, or number of agents change. We evaluate~\proposedmodel~on a point-mass system in two-dimensional space, Kuramoto phase synchronization dynamics and predator-swarm interaction dynamics demonstrating orders of magnitude improvement in prediction accuracy over traditional deep learning models. 
\end{abstract}

\input{introduction.tex}

\input{interaction.tex}
\input{experiments.tex}
\input{results.tex}
\input{conclusion.tex}

%%%%%%%%%%%%%%%%%%%%%%%%%%%%%%%%%%%%%%%%%%%%%%%%%%%%%%%%%%%%%%%%%%%%%%%%%%%%%%%%

\addtolength{\textheight}{-12cm}   % 

%%%%%%%%%%%%%%%%%%%%%%%%%%%%%%%%%%%%%%%%%%%%%%%%%%%%%%%%%%%%%%%%%%%%%%%%%%%%%%%%

%%%%%%%%%%%%%%%%%%%%%%%%%%%%%%%%%%%%%%%%%%%%%%%%%%%%%%%%%%%%%%%%%%%%%%%%%%%%%%%%

%%%%%%%%%%%%%%%%%%%%%%%%%%%%%%%%%%%%%%%%%%%%%%%%%%%%%%%%%%%%%%%%%%%%%%%%%%%%%%%%

%%%%%%%%%%%%%%%%%%%%%%%%%%%%%%%%%%%%%%%%%%%%%%%%%%%%%%%%%%%%%%%%%%%%%%%%%%%%%%%%

\bibliographystyle{IEEEtran}
\bibliography{ref}

\end{document}

%% file: introduction.tex
\section{Introduction}
\label{sec:intro}

Multi-agent systems are prevalent in both the natural world and engineered world. Engineered distributed systems of mobile robots, multiple sensors, unmanned aerial vehicles etc. often take inspiration from natural multi-agent systems like swarms, schools, flocks, and herds of social animals or birds. Understanding the behavior of  such natural or engineered multi-agent systems from sensory observations is a key challenge in robotics from the design and adversarial perspective. 
Discovering the hidden dynamics of a multi-agent interaction from observations will enable machines to simulate and predict evolution of complex systems. 

Research in the field of data-driven dynamics learning can be divided into two main categories. First, one assumes well-known equations of the physical system and estimate their parameters based on observation data \cite{salzmann2011physically, brubaker2009estimating, wu2015galileo, wu2016physics}. However, many complex systems are difficult to represent solely by a fixed model. The alternative (and arguably more compelling) approach is to identify an approximate representation of the actual model using machine learning techniques like regression \cite{brunton} or neural networks \cite{mottaghi2016newtonian, battaglia2016interaction, watters2017visual, chang2016compositional}.
As an important step in this direction, Battaglia et al. \cite{battaglia2016interaction} presented  interaction networks (INs) to learn multi-agent interaction by coupling machine learning with structured models. Watters et al. \cite{watters2017visual} improved IN to learn multi-agent interactions from visual observations. However, IN requires object relation graph as an explicit input; but the relation graphs are often unknown in a real scenario. Moreover, input state vector to IN can include physical properties like agent's mass which may not be directly observable. Chang et al. \cite{chang2016compositional} proposed a similar model to predict bouncing ball dynamics. Their model does not require object relation graph as input and can predict mass of the involved agents; however, they did not demonstrate its ability to predict evolution of dynamics with pairwise interaction force among agents. Finally, these models \cite{battaglia2016interaction,chang2016compositional} are generalized to any number of agents only when physical properties of agents and pairwise interaction parameters remain uniform or explicitly given as input and do not allow online learning or re-tuning with less data in similar scenarios with different physical properties and different interaction parameters.

\begin{figure}[t]
  \centering
  \includegraphics[width=1\linewidth]{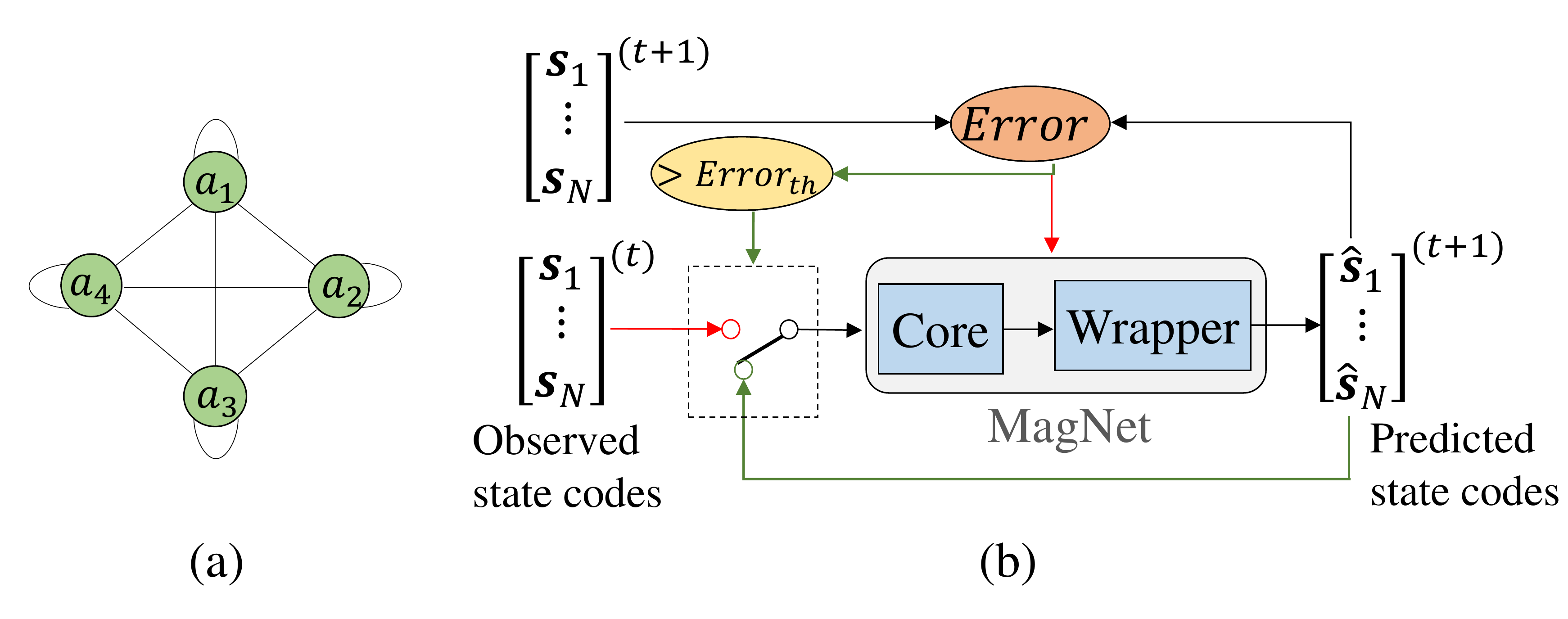}
  \caption{(a): Multi-agent network with four agents. State-dynamics of each agent is dependent on itself and other agents. (b): Training, online re-tuning and prediction mode of ~\proposedmodel. Black arrows belong to all three modes. Red arrows are activated in training and re-tuning mode, whereas green arrows operate only in prediction mode.}
  \vspace{-4mm}
  \label{fig:intro}
\end{figure}

In this paper, we introduce the \proposedmodel (\textbf{M}ulti-\textbf{ag}ent interaction \textbf{Net}work) that can discover interaction dynamics and predict evolution of complex multi-agent system with heterogeneous relational attributes and physical properties \textit{\textbf{solely}} from observational data. The foundation of \proposedmodel~is based on the formulation of multi-agent system as a coupled non-linear network where agents are assumed to be connected to each other using a generic ordinary differential equation (ODE) based state evolution dynamics. The formulation is inspired by a wide range of multi-agent systems ranging from objects interacting by virtue of fundamental laws of physics to swarm systems, opinion dynamics under social interaction \cite{couzin, concensus1, consensus2, yu2013swarming}. \proposedmodel~discovers the dynamics of a multi-agent system by learning the ``customization" of the generic ODE to minimize the error between prediction and sensory observation. ~\proposedmodel~does not require relational graph or non-observable parameters as input, rather it is inherently capable of learning relationship among agents from observations and due to the preceding formulation, agent-specific parameters of the ``customization" can be learned online. The paper makes following key contributions in discovering multi-agent dynamics from observations.

\vspace{-1mm}
\begin{itemize}
   \item We develop a neural network based realization of the time-discretized model of the coupled non-linear network representing multi-agent dynamics that can be trained using stochastic gradient descent (SGD) based backpropagation. The model is trained for single time-step prediction; long term prediction is performed through iterative single-step prediction.
    
    \item The~\proposedmodel~supports continuous learning to accurately predict state evolution even if the relational attributes (e.g. interaction coefficients among agents), physical properties of agents (e.g. mass), or the number of agents changes, but the fundamental interaction remains the same. 
    This is enabled by structuring~\proposedmodel~as two back-to-back networks: a core network to model/learn the fundamental multi-agent dynamics, and a reduced-complexity wrapper network to learn the agent-specific parameters. The entire network is first trained as a single entity. During operation, core network is kept frozen, but the wrapper network is re-tuned once the prediction error crosses a threshold (Figure \ref{fig:intro}(b)).
    
    \item We demonstrate application of~\proposedmodel~for learning/predicting dynamics from direct, as well as noisy observations of states. 
\end{itemize}
\vspace{-1mm}

%% file: interaction.tex
\section{\proposedmodel: Foundation and Design}
\label{sec:interaction}

In this section, we describe the design of our multi-agent interaction network from a generalized formulation of multi-agent dynamical systems. The foundation of our model is built upon the following assumptions:
\begin{enumerate}[label=(\roman*)]
   \item The time evolution of states of the underlying multi-agent dynamical system is a function of pairwise interactions and self-dependence.
   \item The core interaction law for all pairs of agents can be represented by a common form, a linear combination of several interaction terms of different degrees acting simultaneously. However, the coefficients of these interaction terms can be different (including zeros) for each pair of agents depending on their relational attributes (e.g. spring constant in spring systems) or physical properties (e.g. mass in case of n-body gravitational system).
\end{enumerate}
\subsection{Mathematical formulation}
On the basis of assumption (i), the generalized model of multi-agent dynamical systems with $N$ agents can be described by the following system of ODE's
\begin{align}
    \frac{d \mathbf{s}_i (t)}{dt} = g_i\big(\mathbf{s}_i (t)\big) + \sum_{j=1}^N f_{ij} \big(\mathbf{s}_i (t), \mathbf{s}_j (t)\big) \nonumber \\
    \quad \forall i \in \{1, 2, ..., N\}
    \label{eqn:general_model}
\end{align}
The vector $\mathbf{s}_i (t) \in \mathbb{R}^d$ denotes the state of $i^{th}$ agent $a_i$ at time $t$. Function $f_{ij}$ describes the interaction effect from agent $j$ to agent $i$ and function $g_i $ represents the dependence on self-state. 

Considering the assumption (ii), interaction functions $f_{ij}$'s can be written as \begin{equation}
    f_{ij} \big(\mathbf{s}_i (t), \mathbf{s}_j (t)\big) = I_{ij} f\big(\mathbf{s}_i (t), \mathbf{s}_j (t)\big) \quad \forall (i,j)
\end{equation}  
where $f$ delineates the core interaction law and  $I_{ij}$'s are agent specific kernels to actuate the effect of interaction. $I_{ji}$ is not necessarily same with $I_{ij}$. For example, in classical mechanics even though the interaction forces between a pair of objects are equal and opposite, the effect of the interaction on an object i.e. acceleration depends on its own mass. Analogous to equal and opposite forces, we assume the core interaction function $f$ is skew-symmetric in nature:
\begin{equation}
    f\big(\mathbf{s}_i (t), \mathbf{s}_j (t)\big) = - f\big(\mathbf{s}_j (t), \mathbf{s}_i (t)\big) \quad \forall (i,j)
\end{equation}
Skew-symmetric interaction function is modeled as an odd function of inter-agent state difference in a broad set of multi-agent systems ranging from objects interacting by virtue of fundamental laws of physics to swarm systems, opinion dynamics under social interaction \cite{couzin, concensus1, consensus2, yu2013swarming}. Accordingly, our core interaction function $f$ is represented by the following equation. 
\begin{equation}
    f\big(\mathbf{s}_i (t), \mathbf{s}_j (t)\big) = f\big(h(\mathbf{s}_i (t)) - h(\mathbf{s}_j (t))\big)
    \label{eqn:interaction_def}
\end{equation}
Function $h$ models an encoded state of agents. Definition of $f$ in equation \ref{eqn:interaction_def} follows the skew-symmetric property if $f$ is an odd function. Considering all the aforementioned assumptions, our multi-agent interaction model can be delineated by the following system of ODE's 
\begin{align}
    \frac{d \mathbf{s}_i (t)}{dt} = g_i\big(\mathbf{s}_i (t)\big) + \sum_{j=1}^N I_{ij} f\big(h(\mathbf{s}_i (t)) - h(\mathbf{s}_j (t))\big) \nonumber \\
    \quad \forall i \in \{1, 2, ..., N\}
    \label{eqn:our_model}
\end{align}
In this work, our goal is to learn to approximate $f$, $h$, $I$ and $g$ from observable states of agents. Observation data can be contaminated with noise and differentiation of such data, as required by equation \ref{eqn:our_model}, will amplify the noise and therefore, not suitable as target variable during training. To avoid differentiation, we convert the model as iterative update scheme using Eular discretization: 
\begin{align}
    \mathbf{s}_i^{t+1} = \mathbf{s}_i^t + \Delta t \Big( g_i\big(\mathbf{s}_i^t\big) + \sum_{j=1}^N I_{ij} f\big(h(\mathbf{s}_i ^t) - h(\mathbf{s}_j^t)\big) \Big) \nonumber \\
    \quad \forall i \in \{1, 2, ..., N\}
    \label{eqn:discrete_model}
\end{align}
$\Delta t$ is the sampling period of observation. Discretized model enables state to state training without computing derivatives of state vectors. 

\subsection{Implementation with neural networks}
\begin{figure}
  \centering
  \includegraphics[width=1\linewidth]{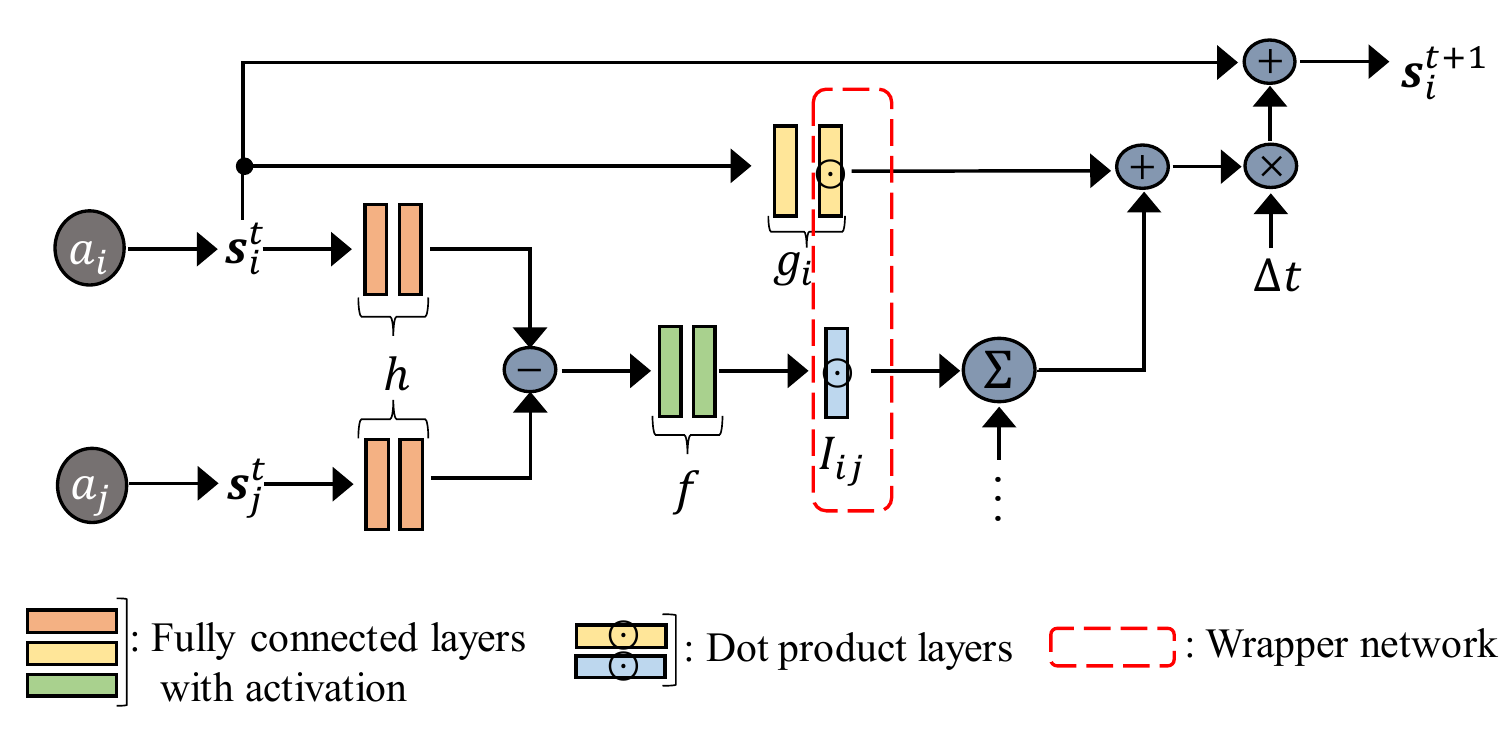}
  \caption{Architecture of ~\proposedmodel~. Layers outside the red dotted box constitute the core network which captures the fundamental laws present in the system. Wrapper network (layers inside the red-dotted box) learns agent specific parameters}
  \vspace{-4mm}
  \label{fig:interaction_model}
\end{figure}

In order to learn the evolution of the dynamical system defined in equation \ref{eqn:discrete_model}, we implement the component functions using standard neural networks and use stochastic gradient descent optimization to train those. Figure \ref{fig:interaction_model} shows the neural network implementation of the discretized multi-agent dynamical system defined in equation \ref{eqn:discrete_model}. 

Each of the functions ($f, g$ and $h$) is implemented with a two-layer fully connected network. All layers of $f$ and $h$, and the first layer of $g$ form the core of the network. Weights of these core layers are shared across all agents and are independent of number of agents present in the system. Core layers are responsible for modeling the fundamental interaction laws and self-dependence. Number of layers and number of neurons in each layers should be customized based on the expected degree of non-linearity in the system.  

Weight matrix $I_{ij}$ and second layer of function $g$ are agent-specific and work as a wrapper network on top of the core network. Wrapper network is responsible for the physical properties of the agents (i.e. interaction coefficients, mass etc.). Wrapper network scales with number of agents so as the available data for updating corresponding weights online. To reduce the number of weights per agents, we use dot-product layers instead of fully-connected layers. Suppose, the length of the feature vector out of the function $f$ is $L$ and $d$ is the length of the agent's state code. We choose $L$ such that $L = ld$, where $l$ is an integer. Now, each component of length $l$ from the feature vector contribute to only one component of the state code. 
Hidden feature vector of length $L$ is reshaped as a matrix of size $l \times d$ before feeding it to the dot-product layer. Operation of the dot-product layer is defined as follows
\begin{equation}
    [\mathbf{e}_1 \quad \cdots \quad \mathbf{e}_d] \odot [\boldsymbol{\omega}_1 \quad \cdots \quad \boldsymbol{\omega}_d]  = [\mathbf{e}_1^T \boldsymbol{\omega}_1 \quad \cdots \quad \mathbf{e}_d^T \boldsymbol{\omega}_d],
\end{equation}
where $\mathbf{e}_k \in \mathbb{R}^l$ and $\boldsymbol{\omega}_k \in \mathbb{R}^l$. 

Any nonlinear activation function can be used for function $h$ and the first layer of function $g$. We use rectified linear units (ReLUs) for these layers. In order to hold the skew-symmetric property, an odd activation function is required for the layers of $f$. We use $tanh$ for this purpose. For the same reason, the layers of $f$ are implemented as linear transform without adding any bias.  

%% file: experiments.tex
\section{Experimental Details}
\label{sec:experiments}

\subsection{Datasets}
We consider three different multi-agent dynamics to demonstrate the performance of ~\proposedmodel.

\textbf{Point-mass system} 
Agents in this dataset are objects with different mass moving in a two-dimensional space according to Newton's laws of motion. 
We consider two types of forces are acting simultaneously between each pair of agents. The first interaction force is due to invisible spring between each pair of agents. We consider different spring constants for different pairs. 
The second kind of force is a repulsive inverse square law force between each pair. This force is proportional to the product of mass of the involved agent-pair. 
Pairwise-interaction for the considered dynamics is given by the following equation
\begin{equation}
    \mathbf{F}_{ij} = -k_{ij}(\mathbf{p}_i - \mathbf{p}_j) + K \frac{m_i m_j (\mathbf{p}_i - \mathbf{p}_j)}{(\lambda + \Vert (\mathbf{p}_i - \mathbf{p}_j) \Vert )^3} ,
    \label{eqn:interaction_force}
\end{equation}
where $\mathbf{p}_i \in \mathbb{R}^2$ is the position of the $i^{th}$ agent and $m_i$ is its mass. $\mathbf{F}_{ij} \in \mathbb{R}^2$ is the force agent $j$ exerts on agent $i$, $k_{ij} > 0$ is the spring constant for agent-pair $(i,j)$, $K >0$ is the coefficient for repulsive force and $\lambda > 0$ is some constant to clip the repulsive force to a finite value when two agents are very close. We use $\lambda = 10$.

\textbf{The Kuramoto model} This is a well-known non-linear dynamical model used to described the synchronization of a set of coupled oscillators. Behavior of many biological and chemical oscillators can be described by this model \cite{acebron2005kuramoto}. Each oscillator tries to run independently at its own natural frequency, while the coupling tends to synchronize it to others. Dynamics of $i^{th}$ oscillator is given by
\begin{equation}
    \frac{d \theta_i}{dt} = \omega_i + \sum_{j=1}^N K_{ij} \  sin(\theta_j -\theta_i),
\end{equation}
where $\theta_i$ and $\omega_i$ are the phase and natural frequency, respectively, of the $i^{th}$ oscillator. $N$ is the number of oscillators in the system. $K_{ij}$ is the coupling coefficient between oscillator-pair $(i,j)$. 

\textbf{Predator-swarm interaction dynamics}
This dynamics is similar to the one used to describe the behavior of prey swarm in presence of predators \cite{chen2014minimal}. Dynamics of the system with $N$ prey and one predator is given by the following set of equations:
\begin{align}
    \frac{d \mathbf{x}_i}{dt} &= \frac{1}{N} \sum_{j=1}^N \bigg(\frac{\mathbf{x}_i - \mathbf{x}_j}{| \mathbf{x}_i -\mathbf{x}_j|^2} - a(\mathbf{x}_i -\mathbf{x}_j) \bigg) + b \frac{\mathbf{x}_i - \mathbf{z}}{| \mathbf{x}_i -\mathbf{z}|^2} \nonumber \\
    \frac{d \mathbf{z}}{dt} &= \frac{c}{N} \sum_{j=1}^N \frac{\mathbf{x}_i - \mathbf{z}}{| \mathbf{x}_i -\mathbf{z}|^2},
\end{align}
where $\mathbf{x}_i$ denotes the position of $i^{th}$ prey and $\mathbf{z}$ denotes the position of the predator. 

Data for all systems is generated using finite difference method with small timestep. Sequences for training, validation and testing are created by choosing initial states randomly. 

\subsection{Implementation details}
For our point-mass dataset, state code of agents is the concatenated position and velocity components along both dimensions ($\mathbf{s}_i \in \mathbb{R}^4$). We predict the acceleration vector of length 2 for each agent. Velocity vector for next state is not predicted by the network directly, rather we compute it from acceleration and current velocity. Finally, next position is computed from the current position and predicted velocity. Number of neurons in both layers of function $h$ is 64. The first layer of function $f$ consists of 64 neurons while second layer has 8 neurons. Therefore, the output of function $f$ is length 8 vector which is reshaped in to a matrix of size $4 \times 2$ for the following dot product layer. $I_{ij}$'s are matrices of size $4 \times 2$. First layer of functions $g_i$'s are of size 4 and are shared among all agents. Outputs from the first layers of $g_i$'s are reshaped in to matrices of size $2 \times 2$ for the following dot product layers (one for each agent). We also add agent-wise bias in these dot product layers. Table \ref{tab:flop} shows the total number of parameters and FLOP count of the used network for $N$ agents.  

Same implementation is used for predator-swarm interaction dynamics and the Kuramoto model except the changes required for state code dimension. For Kuramoto model, phase of the oscillating agents are used as the state code ($\mathbf{s}_i \in \mathbb{R}$). 

\begin{table}[h]
  \caption{Parameter count and FLOP count of ~\proposedmodel~ used in our experiments for $N$ agents}
  \label{tab:flop}
  \centering
  \begin{tabular}{ccc}
    \toprule
         & Parameter count  & FLOP count    \\
    \midrule
    Core network & $9108$ & $17880N$\\
    Wrapper network & $8N^2-2N$ & $14N^2$ - $6N$ \\
    \bottomrule
  \end{tabular}
  \vspace{-4mm}
\end{table}

\subsection{Baseline models}
We consider the following baseline models to compare accuracy of~\proposedmodel. 

\textbf{Linear motion} Linear motion model assumes the velocity of the state is constant. We compute the velocity of state from previous two timesteps and predict the next state using first order approximation.

\textbf{MLP} We use a baseline MLP that takes the concatenated state codes from all agents as input and predict the same for next timestep. This configuration does not share any weights among agents and therefore, is not scalable with number of agents. For four-agent system, we use three hidden layers, each of size 64, followed by two layers of size $N \times dim(\mathbf{s}_i)$, where $dim(\mathbf{s}_i)$ denotes the dimension of vector $\mathbf{s}_i$. Size of the network is chosen to have similar parameter count with ~\proposedmodel.

\textbf{LSTM} We use a baseline LSTM that uses state codes from previous four timesteps to predict the next state. Similar to baseline MLP, the LSTM model does not share any weights among agents and therefore, is not scalable with number of agents. For four-agent system, we use a two-layer LSTM (each layer is of size 64). The LSTM core is preceded by a linear layer of size 64 and is followed by a output linear layer of size $N \times dim(\mathbf{s}_i)$.

\subsection{Training and online re-tuning}
~\proposedmodel~ is trained or re-tuned as a single-step predictor from current state to next state with $M \times L$ number of observations. $M$ denotes the number of random initial conditions and $L$ denotes the length of each sequence generated from those initial conditions. $SmoothL1$-loss is used as the objective function. State variables are standardized to have zero mean and unit variance. We use Adam optimizer \cite{adam} to optimize the parameters.  

We consider two training scenarios for point-mass dynamics. In the first case, we assume perfect observation data (no noise). The second case considers observation data contaminated with Gaussian noise. Core network and wrapper network are trained together with $M=50$ and $L=500$. We train the model for $100$ epochs starting with an initial learning rate of $10^{-3}$ and scaled it by a factor of 0.95 after each epoch until it reaches $10^{-4}$. Differentiating noisy position vectors of agents to compute their velocities amplifies the noise in velocity vectors. We use total variation regularization \cite{TV} to denoise the derivatives \cite{TVdiff} as suggested in \cite{brunton}.

In online re-tuning, we cannot have multiple random initial condition. Therefore, value of $M$ must be equal to 1 while value of $L$ should be much larger (we use $L=10000$) to avoid overfitting. We start with an initial learning rate of $5 \times 10^{-4}$ and scaled it by a factor of 0.95 after each epoch.

For Kuramoto model, we use $8$ oscillators with different intrinsic frequencies and different pairwise coupling coefficients. We use the same training setting and same amount of data (i.e. $M=50, L=500$) as used in point-mass dynamics. In predator-swarm interaction, we use $20$ prey in presence of one predator and the model is trained using $M=100$ and $L=300$ for $100$ epochs with constant learning rate of $10^{-4}$.

We  considered  tuning few  hyperparameters like changing the number of neurons in hidden layers in powers of 2,  learning rates in range from $5 \times 10^{-5}$ to  $5 \times 10^{-3}$. Number of neurons in hidden layers are selected such that parameter count is not too high and accuracy is reasonable as well. We found the chosen learning rate schedule works well towards reaching convergence. 

\let\thefootnote\relax
\footnotetext{\scriptsize Code and demo videos are available at \href{https://github.com/sahapriyabrata/MagNet}{https://github.com/sahapriyabrata/MagNet}}

%% file: results.tex
\section{Results}
\label{sec:results}

\begin{figure*}[t]
  \centering
  \includegraphics[width=0.9\linewidth]{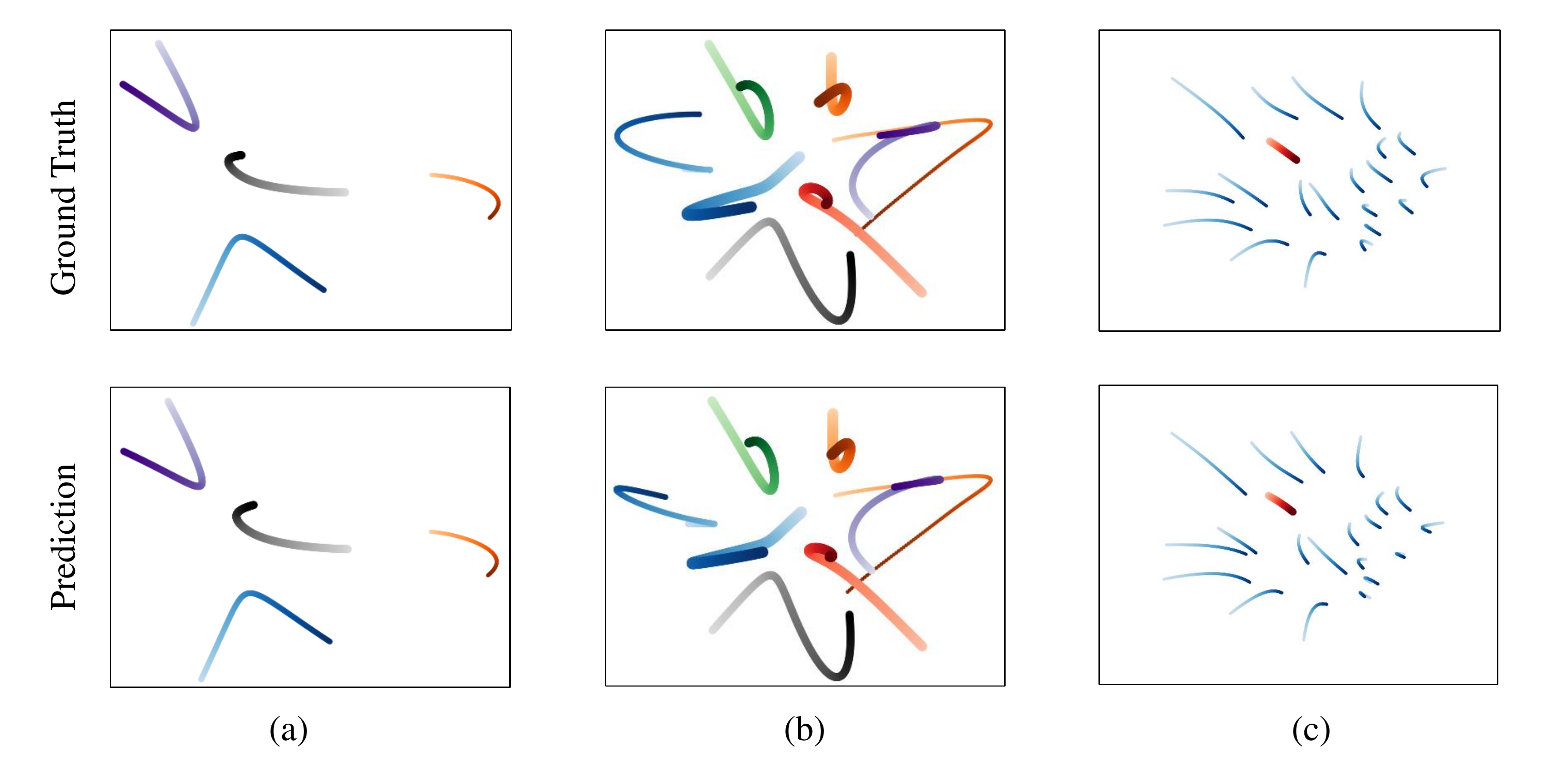}
  \caption{Visualization of evolution up to 200 timesteps. (a) Trajectory plot of point-mass system with four (4) agents. Widths of the trajectories are proportional to the masses of corresponding agents. Predictions are from network trained from scratch. (b): Trajectory plot of point-mass system with eight (8) agents. Predictions are from re-tuned wrapper network preceded by frozen core network trained with 4 agents. (c): Trajectory plot of predator-swarm system with twenty (20) prey and one (1) predator. Red wider trajectory correspond to the predator. Predictions are from network trained from scratch.}
  \label{fig:visual}
\end{figure*}

\begin{figure}
  \centering
  \includegraphics[width=1\linewidth]{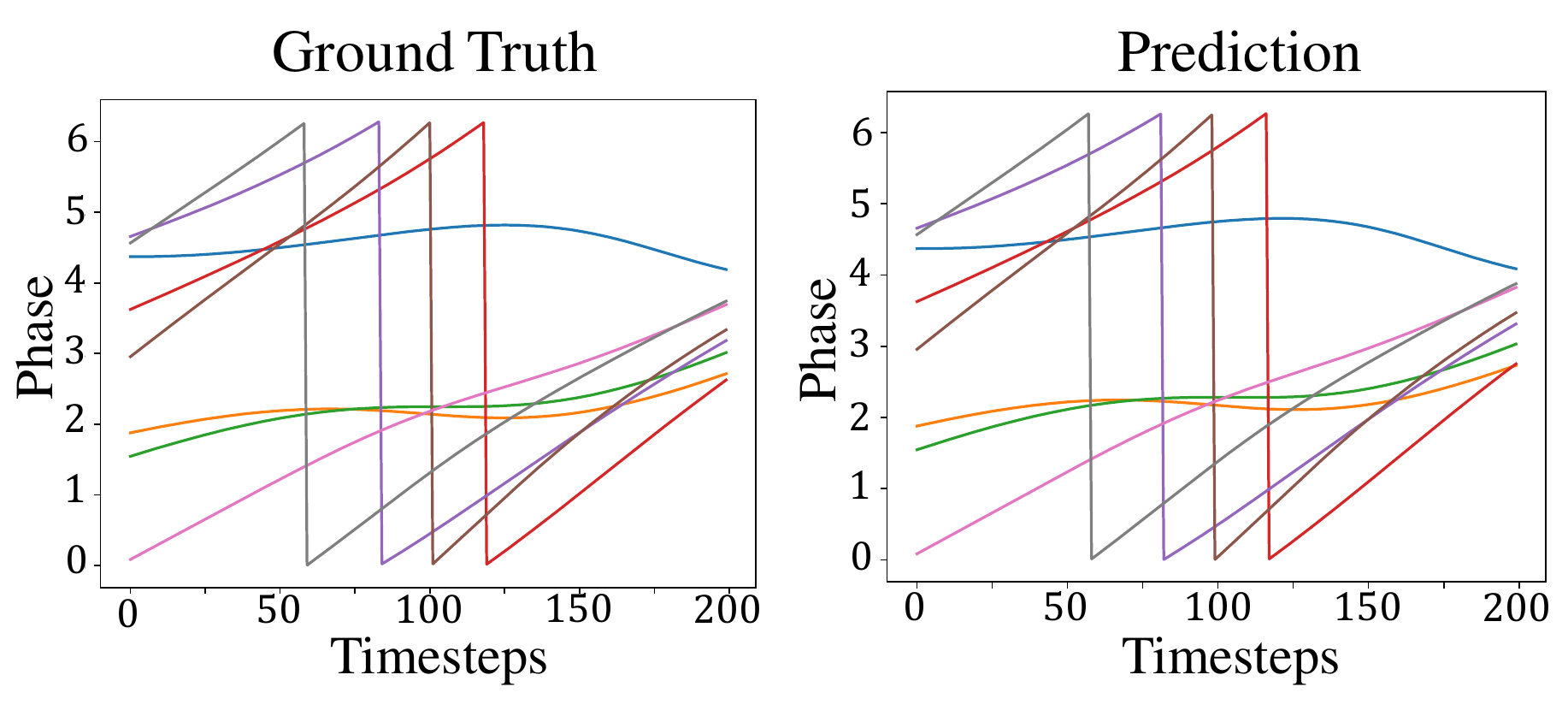}
  \caption{Visualization of evolution up to 200 timesteps. Phases ($0$ to $2\pi$) over timesteps for eight (8) oscillating agents abide by the Kuramoto model. Predictions are from network trained from scratch.}
  \vspace{-2mm}
  \label{fig:visual2}
\end{figure}

All results are generated as solution to an initial value problem i.e. evolution of the system is predicted only from an initial observation, no intermediate observation is used. We use mean-squared-error (MSE) between ground truth and prediction through timesteps as metric for evaluation. Fifty ($50$) test sequences are used to generate the MSE plots with errorbars showing $95\%$ confidence intervals. Visual evolution of ground truth and prediction are shown in Figure \ref{fig:visual} and Figure \ref{fig:visual2}.

\begin{figure*}[t]
  \centering
  \includegraphics[width=0.8\linewidth]{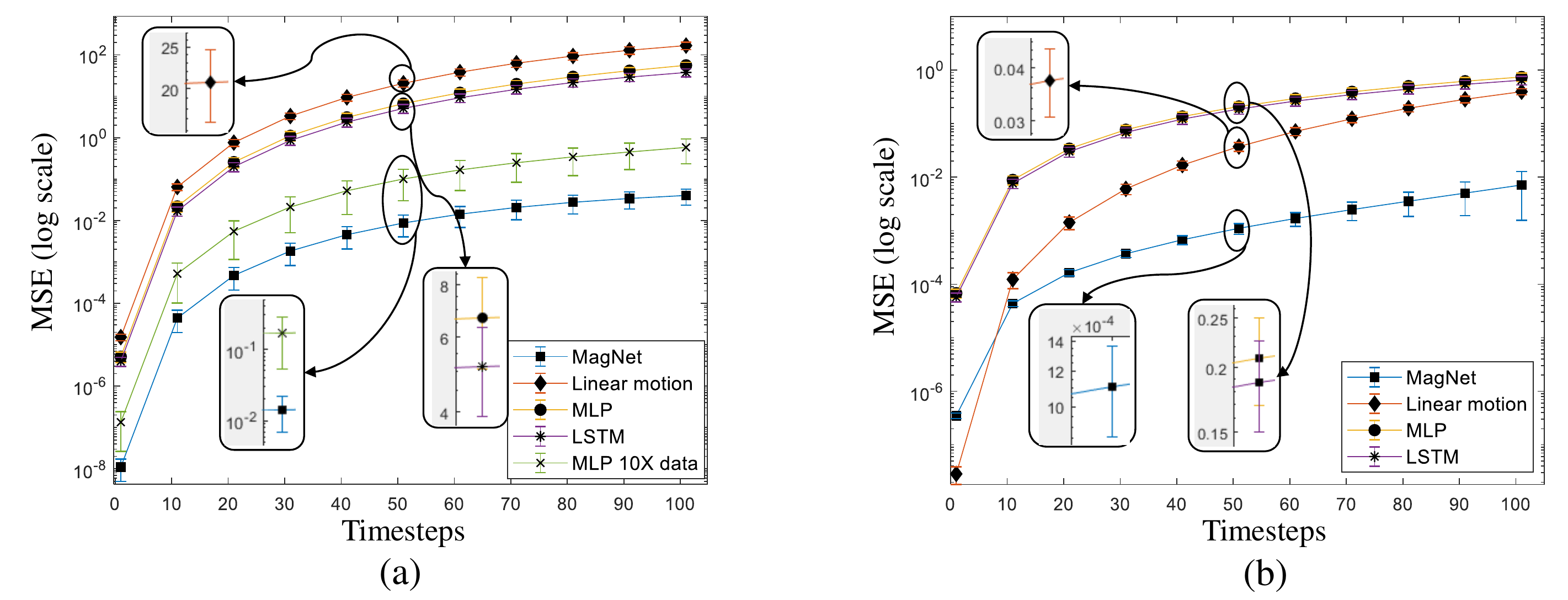}
  \caption{(a): MSE between ground truth positions and predicted positions of agents of considered point-mass system. `MLP' denotes MLP trained with same amount of data as MagNet, whereas `MLP 10X data' denotes MLP trained with 10X more data and 10X more steps. (b): MSE between ground truth phases and predicted phases of oscillating agents in Kuramoto model.}
  \label{fig:main_train}
\end{figure*}

\begin{figure*}[t]
  \centering
  \includegraphics[width=1\linewidth]{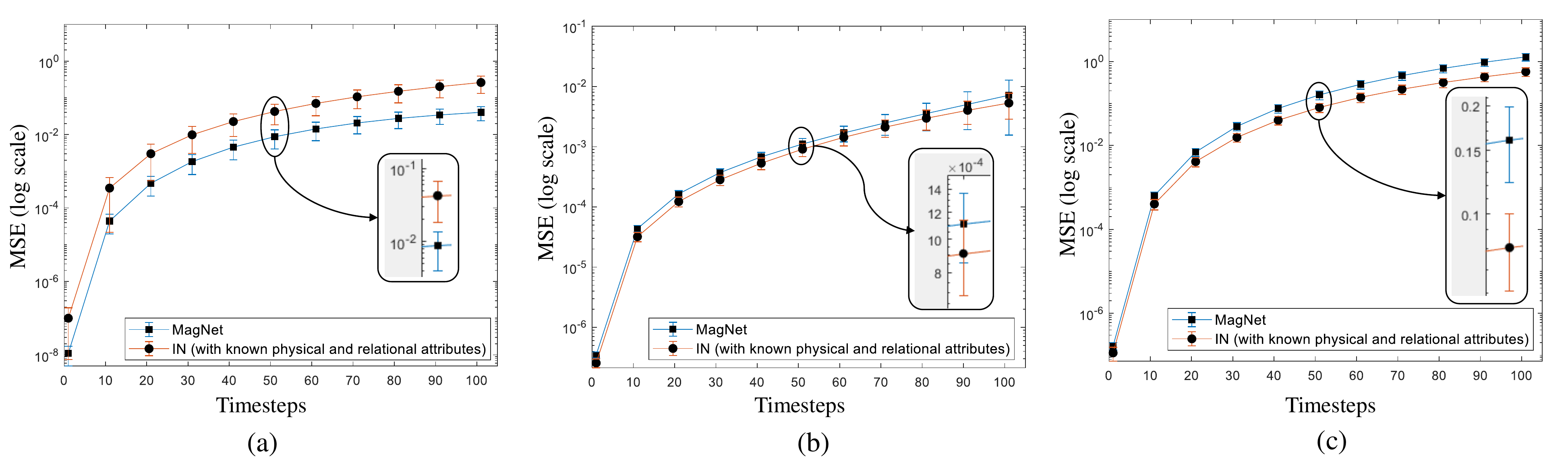}
  \caption{Comparison with IN, which takes physical and relational attributes as input. (a): MSE between ground truth positions and predicted positions of agents of considered point-mass system. (b): MSE between ground truth phases and predicted phases of oscillating agents in Kuramoto model. (c): MSE between ground truth positions and predicted positions of agents of considered predator-swarm system. }
  \label{fig:withIN}
\end{figure*}

\subsection{Learning and prediction from direct and clean observations} We consider four ($4$) interacting objects with different mass and different pairwise spring constants for point-mass system. ~\proposedmodel~ can predict the evolution of state codes for a long period of time with negligible error if it is trained with perfect observations (no noise). Figure \ref{fig:main_train} shows the MSE between ground truth and prediction over timesteps for MagNet along with all baselines for point-mass system and Kuramoto model. 
As shown in Figure \ref{fig:main_train}(a), even if the baseline MLP is trained with more data (we use 10X more data and 10X more number of steps than~\proposedmodel), the MSE is higher than ~\proposedmodel. Note, the baseline MLP is not scalable with number of agents; hence, data requirement would increase exponentially with number of agents. Accordingly, training MLP or LSTM baseline for predator-swarm dynamics with twenty-one (21) agents is intractable and hence, is not considered for comparison. 

\subsection{Comparison with interaction network \cite{battaglia2016interaction}} IN \cite{battaglia2016interaction} requires physical and relational attributes of the agents as input along with their observable states. Therefore, IN is trained and evaluated assuming the physical and relational attributes of agents are known. In contrast, our model is trained and evaluated using only the observable states. Size of the implemented IN is chosen to have similar parameter count with our model. Figure \ref{fig:withIN} shows the performance comparison between our model an IN. Our model shows comparable performance (better for point-mass system) with IN, which has access to physical and relational attributes of agents.

\subsection{Learning and prediction from noisy observations} 
\begin{figure*}[t]
  \centering
  \includegraphics[width=0.8\linewidth]{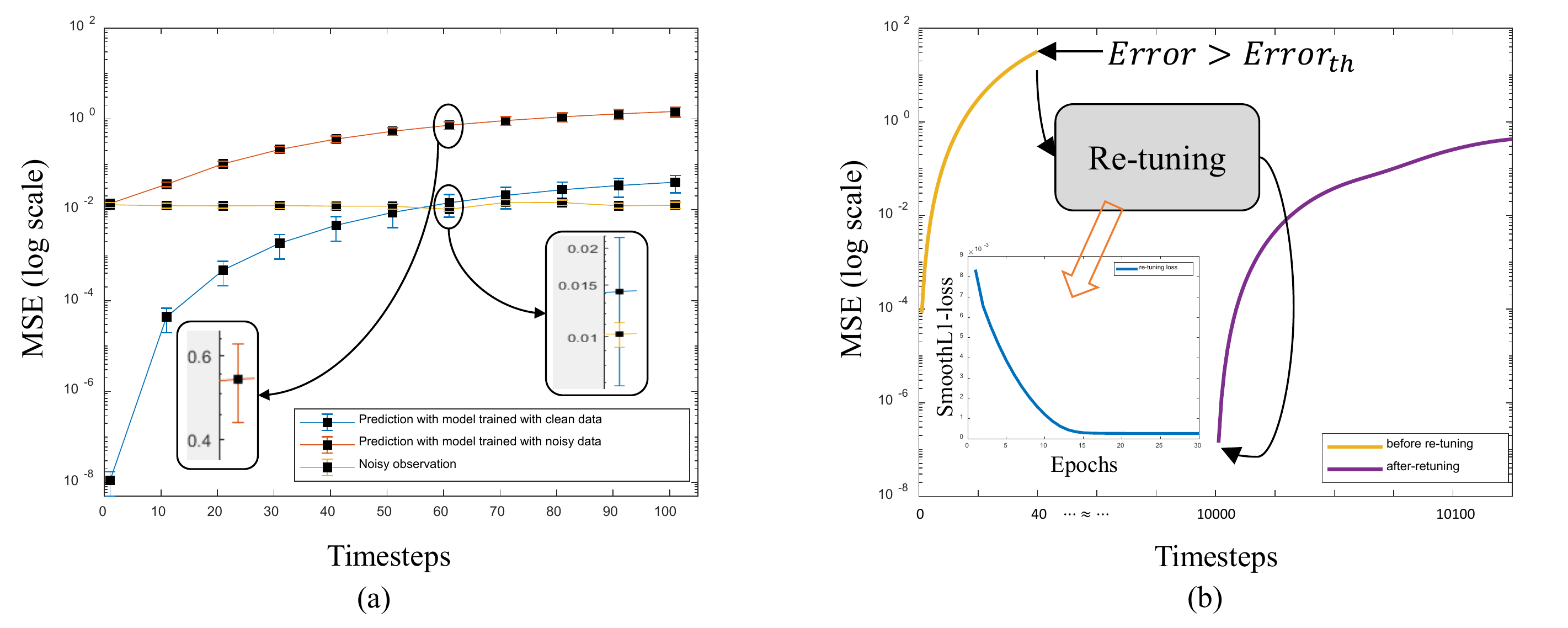}
  \caption{(a): MSE with ground truth positions for~\proposedmodel~trained with noisy observation of point-mass system. (b): MSE before and after re-tuning in a scenario with different number of agents from training scenarios. Re-tuning loss is shown in the inset.}
  \vspace{-2mm}
  \label{fig:main_tune}
\end{figure*}

While evaluating the model on test sequences, we use initial 16 observations to denoise the derivatives (velocities) using total-variation regularization \cite{TV, TVdiff}. Figure \ref{fig:main_tune}(a) shows the MSE over timesteps for the model trained with noisy observation. As expected, when dynamics is learned from noisy observations, accurate prediction window becomes shorter than that of with perfect observation. However, we observe that MSE of the network trained with noisy observation remains within 10X margin of the network trained with clean observation up to 100 timesteps.

\subsection{Performance of re-tuning}
In this experiment, we increase the number of agents for the point-mass system to eight (8) and change spring constants between agent-pairs and masses of the agents. We seek to predict evolution of this eight-agent system using the~\proposedmodel~trained with four (4) agents. Agent-wise wrapper-weights are initialized with the average values of pre-trained wrapper-weights across all agents. Figure \ref{fig:main_tune}(b) shows that the prediction error increases with time and once crosses a threshold, re-tuning of the wrapper (core is kept frozen) starts. We observe that after re-tuning with 10000 observations, prediction error for the eight-agent system reduces (Figure \ref{fig:main_tune}(b)). This experiment demonstrates the generalization capability of the core network within~\proposedmodel~.

%% file: conclusion.tex
\section{Conclusion}
\label{sec:conclude}
We introduced the~\proposedmodel~ to discover multi-agent dynamics from sensory observations. We showed that the proposed model can identify the inherent dynamics and predict its evolution. 
We observe that a major advantage of~\proposedmodel~over state-of-the-art is that it can be re-tuned online if the relation parameters or physical properties of agents get altered or the number of agents is changed, but the fundamental laws remain same. This capability makes~\proposedmodel~employable in real scenarios where these relation parameters and physical properties often change and may not be directly observable.

One limitation of the current model is that it weights different interaction terms in a linear way with relational attributes or physical parameters. This assumption may not be true in many cases. In future, we would to like to address this shortcoming. Moreover, we plan to extend~\proposedmodel~such that on-line tuning can be performed to reduce error even when the core dynamics is changed over time. Exploring~\proposedmodel~to learn dynamics of agents controlled by external input to achieve some goals will be an important extension as well.